\documentclass[sigconf,natbib=true,anonymous=false]{acmart}
\usepackage{xcolor}
\usepackage{booktabs}
\usepackage{soul}
\usepackage{tabularx}
\usepackage{multirow}
\usepackage{makecell}
\usepackage{adjustbox}
\usepackage{microtype}
\usepackage{amsmath}

\usepackage{array}
\usepackage{pgfplots}
\usepackage{tikz}
\usetikzlibrary{positioning,calc}
\pgfplotsset{compat=1.18}
\pgfplotsset{
  dotpanel/.style={
    xbar,
    ytick align=outside, xtick align=outside,
    axis line style={very thin, color=black!30},
    tick style={very thin, color=black!30},
    ticklabel style={font=\scriptsize},
    ymajorgrids=false, xmajorgrids=true,
    grid style={very thin, color=black!10, densely dotted},
    enlarge y limits=0.04,
    enlarge x limits={upper, value=0.05},
    every axis plot/.append style={line width=0pt},
  }
}

\AtBeginDocument{%
  }

\begin{document}

\title{TRACE: \textbf{T}rustworthy \textbf{R}etrieval-\textbf{A}ugmented \textbf{C}onversational \textbf{E}ngine}

\author{Touseef Hasan}
\affiliation{%
  \institution{Wichita State University}
  \city{Wichita}
  \state{KS}
  \country{USA}
}

\author{Laila Cure}
\affiliation{%
  \institution{Wichita State University}
  \city{Wichita}
  \state{KS}
  \country{USA}
}

\author{Souvika Sarkar}
\affiliation{%
  \institution{Wichita State University}
  \city{Wichita}
  \state{KS}
  \country{USA}
}

\renewcommand{\shortauthors}{Hasan et al.}

\begin{abstract}
Public service chatbots are expected to deliver recommendations from an underlying public service directory, while also making sure that the recommendations respect explicit user constraints. In practice, public service directories are noisy and inconsistent, and general-purpose large language model (LLM) or AI-based chatbots frequently generate unreliable recommendations, citing unverified sources from the web. We investigate the impact of retrieval quality on constraint-aware recommendation in public service conversational systems built over noisy and heterogeneous service directories. We propose \textbf{TRACE} (\textbf{T}rustworthy \textbf{R}etrieval-\textbf{A}ugmented \textbf{C}onversational \textbf{E}ngine), a retrieval-based, constraint-aware framework that parses input user queries into structural and semantic constraints for downstream retrieval, with the help of a dual data representation schema. Using a curated statewide pantry directory and a synthetic query benchmark, we evaluate multiple knowledge-representation variants with and without knowledge graphs (KGs). We experiment with several open-source LLMs and a proprietary model, showing that strengthening retrieval substantially improves user constraint satisfaction while reducing hallucinated recommendations. Performance differences across LLMs narrowed in our experiments as retrieval quality improved, making results less sensitive to model size. These findings suggest that the quality of retrieval is key for robust public service conversational systems.
\end{abstract}

\keywords{Information retrieval, Large language models, Knowledge graphs, Retrieval-augmented generation, Public service recommendation.}


\maketitle

\definecolor{smallc}{RGB}{190,190,190}
\definecolor{mediumc}{RGB}{130,130,130}
\definecolor{largec}{RGB}{70,70,70}
\definecolor{propc}{RGB}{0,0,0}

\definecolor{kg0}{HTML}{C8DCE8}   
\definecolor{kg1}{HTML}{7AAFC8}   
\definecolor{kg2}{HTML}{2E7EA6}   
\definecolor{kg3}{HTML}{0D3D56}   

\section{Introduction}

Public service conversational systems are typically directory-grounded systems, i.e., they answer user requests by retrieving candidate providers from an underlying public service directory~\cite{dow2018between}. These chatbots are expected to deliver reliable recommendations (e.g., where to obtain food assistance) from the directory while respecting explicit user constraints such as location, operating hours, and eligibility requirements~\cite{larsen2024impact, van2019new}. In practice, these constraints are difficult to satisfy because public service directories are often noisy and inconsistently formatted ~\cite{dow2018between, cherry2002uses}. As a result, general-purpose large language model (LLM) or AI-powered chatbots may generate unreliable recommendations by citing unverified sources or implicitly violating user constraints~\cite{dreyling2024challenges}. In preliminary trials with marketplace LLM chatbots (e.g., ChatGPT), out-of-directory recommendations and occasional reliance on unverified web sources were observed, particularly under explicit constraints~\cite{kleiman2025management}. These failures are especially costly in community-help settings, where users rely on the system for reliable time-sensitive guidance.

\noindent
These failures point to a fundamental gap: without reliable retrieval over an authoritative directory, even capable LLMs cannot deliver trustworthy recommendations in public service settings. Our premise is that \emph{generation quality in this domain is largely determined by retrieval quality}: if the system can reliably retrieve a small set of candidates that satisfy the user’s constraints, the LLM’s role is primarily to synthesize a response from the evidence than to ``invent'' new recommendations. To operationalize this idea, we introduce \textbf{TRACE} (\textbf{T}rustworthy \textbf{R}etrieval-\textbf{A}ugmented \textbf{C}onversational \textbf{E}ngine), a constraint-aware framework that relies on robust retrieval prior to LLM response generation. We evaluate our framework on a real-world food pantry dataset built from a curated directory. We further develop a synthetic benchmark of constraint-driven user queries with ground-truth answers. Our evaluation compares multiple knowledge representation variants (with and without knowledge graphs) and multiple LLMs under an identical retrieval setup, measuring constraint satisfaction and out-of-dataset hallucinations. Results support the view that retrieval quality is the primary lever for dependable public service conversational access. Through TRACE, our key contributions are as follows:

\vspace{-2mm}

\begin{itemize}
\item \textbf{Dual-representation, constraint-aware retrieval framework.} A retrieval-first pipeline that explicitly parses user constraints into: (i) structural constraints via knowledge graphs and (ii) semantic constraints via text. This dual representation grounds LLM response generation in directory records and reduces out-of-directory recommendations.
\item \textbf{Knowledge graph (KG) structure ablation for conversational retrieval.} We systematically compare multiple knowledge graph representations with a no-KG baseline and report retrieval and end-to-end reliability metrics.
\item \textbf{Retrieval-centric robustness across LLMs.} Under a fixed retrieval pipeline, we evaluate 15 open-source LLMs and one proprietary model, showing that stronger retrieval substantially narrows performance differences across LLMs (regardless of size) in the public service directory setting.
\end{itemize}

\section{Related Work}

\begin{figure*}
\centerline{\includegraphics[width=1\textwidth, trim=75 370 150 500, clip]{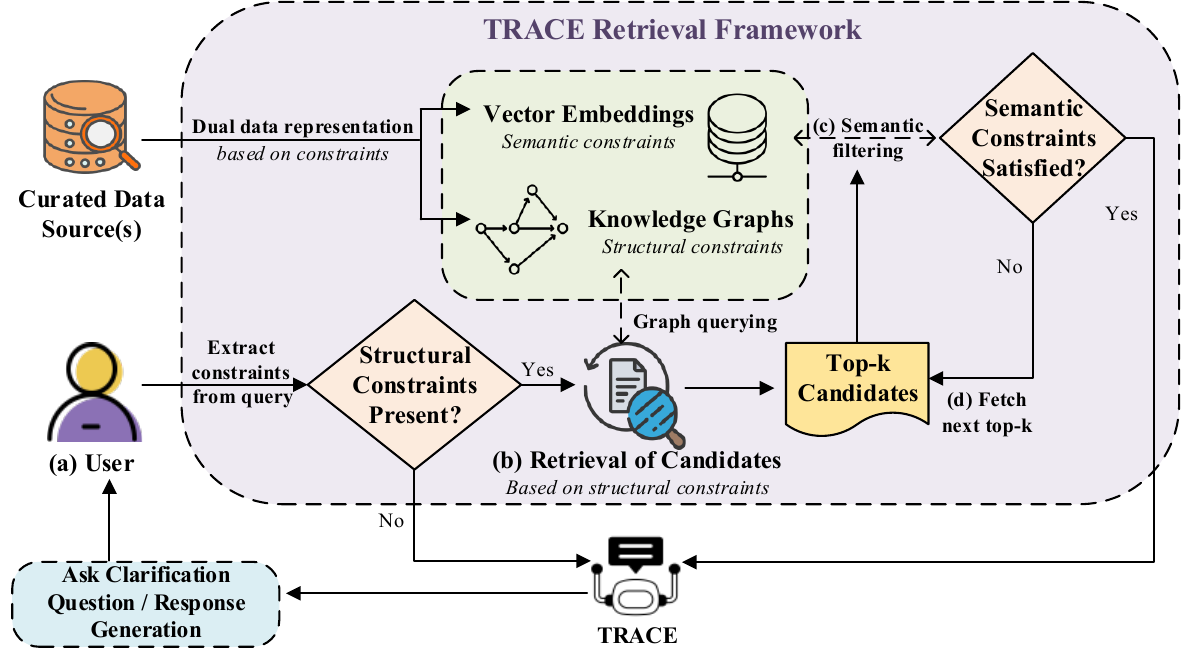}}
    \vspace{-5mm}
    \caption{Overview of the TRACE (Trustworthy Retrieval-Augmented Conversational Engine) framework.}
    \vspace{-2mm}
    \label{fig:framework}
    \vspace{-3mm}
\end{figure*}

\textbf{Constraint-aware conversational systems.} Task-oriented dialogue treats user requests as structured slots (e.g., location, time, eligibility) that must be filled to complete an information need \cite{cohen2019foundations, louvan2020recent}. Recently, the conversational information retrieval (IR) community has emphasized that systems must (i) interpret underspecified queries \cite{malaviya2025contextualized}, (ii) ask clarification questions when key constraints are missing \cite{zamani2020generating, majumder2021ask}, and (iii) maintain constraint consistency across the dialogue \cite{gao2020recent}. Prior work on clarification question generation and conversational search demonstrates the value of explicitly identifying missing facets and eliciting them, rather than guessing or returning loosely related results \cite{tavakoli2020generating, sekulic2021towards}.

\noindent
\textbf{Reliable public-service question answering (QA).} Public-service assistants face a distinctive reliability requirement: answers should be grounded in a verified directory rather than generated from open-web sources \cite{hasan2026retrieval, arends2024chatbot, stamatis2020using}. Document-grounded dialogue benchmarks such as \textsc{doc2dial} \cite{feng2020doc2dial} and \textsc{MultiDoc2Dial} \cite{feng2021multidoc2dial} formalize this setting by requiring systems to retrieve supporting evidence from documents to produce answers. In parallel, recent work has shown that general-purpose LLMs produce fluent but incorrect statements and confidently present unsupported claims, reinforcing the need for explicit grounding and verification \cite{tripathi2025confidence}. These findings are especially relevant in community-help scenarios, where unverified recommendations directly reduce trust and utility \cite{aoki2020experimental, zhou2025improving}. 

\noindent
\textbf{Knowledge graphs (KGs) for retrieval.} KGs have been widely studied to organize entities and relations for search and QA \cite{huang2019knowledge, khan2023knowledge}. In IR, entity-centric representations and graph structures support precise filtering and reduce wrong-entity retrieval \cite{reinanda2020knowledge, dietz2018utilizing}. In QA, hybrid approaches combine graph traversal with text to handle both relational constraints and free-text details \cite{agarwal2025hybrid, polignano2021together}, and recent ``graph-enhanced'' retrieval frameworks use graph structures to guide candidate selection before ranking evidence \cite{li2025simple, yu2026can, zhu2025knowledge, xu2024retrieval}.

\noindent
\textbf{Uniqueness of our work.} While prior work has studied (i) conversational systems that elicit missing slots and constraints, (ii) document-grounded QA to reduce unsupported answers, and (iii) KG representations for retrieval and hybrid QA; they are rarely evaluated together in a public-service directory setting where users pose explicit constraints and the cost of wrong recommendations is high. Our work is distinct in three ways. \textbf{First}, we frame public-service assistance as constraint-aware conversational retrieval and operationalize a dual-constraint pipeline that enforces evidence retrieval and candidate checking. \textbf{Second}, instead of assuming a fixed KG schema, we perform a KG-structure ablation tailored to public-service data, quantifying how representation choices affect retrieval quality. \textbf{Third}, we study the hypothesis that retrieval dominates generation in this setting by evaluating multiple LLMs.

\vspace{-5mm}

\section{Proposed Framework}
\label{sec:framework}

Illustrated in Figure~\ref{fig:framework}, we propose \textbf{TRACE} (\textbf{T}rustworthy \textbf{R}etrieval-\textbf{A}ugmented \textbf{C}onversational \textbf{E}ngine), for public-service conversational search. Given a user query, TRACE (a) extracts structural constraint(s) and, if none are present, asks a clarification question; (b) retrieves a top-$k$ candidate set satisfying the constraints; (c) applies semantic filtering over the retrieved candidates, and (d) if no candidate satisfies the semantic constraints, fetches the next top-$k$ candidates. Once semantic constraints are satisfied, TRACE generates the final response based on the retrieved candidate set.

\vspace{-3mm}

\subsection{Parsing user constraints}
The framework is designed around two types of constraints: (i) structural and (ii) semantic. Given an input user query $q$, we extract:
\begin{itemize}
    \item \textbf{Structural constraints $C_s$}: attributes to be enforced exactly (e.g., city/county/ZIP). These constraints define the candidate space and are evaluated with strict AND semantics. When no structural constraints are present in $q$, the system asks a targeted clarification question (e.g., ``What is your nearby ZIP, city, or county?'') to obtain the structural constraint(s).
    \item \textbf{Semantic constraints $C_m$}: expressed in natural language that typically appear in unstructured text fields (e.g., eligibility requirements). These constraints are checked using retrieved evidence from text fields within the candidate set.
\end{itemize}

\noindent \textbf{Example.} For the query $q$: ``Show me pantries in Sedgwick County that need ID,''
we extract structural constraints $C_s=\{\texttt{county}=\text{Sedgwick County}\}$ and semantic constraints
$C_m=\{\text{need ID}\}$. We first retrieve candidates by querying the KG using $C_s$,
then verify $C_m$ using textual evidence from the candidates' \texttt{eligibility} fields.

\vspace{-3mm}

\subsection{Dual knowledge representation}
We construct a dual representation of the public service directory:
\begin{enumerate}
    \item \textbf{Knowledge graphs (structural view):} Each service provider (e.g., pantry) is linked to a structured entity such as location (city, county, ZIP). We evaluate multiple knowledge graph (KG) variants as ablations against a no-KG baseline.
    \item \textbf{Vector embeddings (semantic view):} For each provider, we retain unstructured fields such as eligibility requirements. These fields are embedded to support semantic retrieval.
\end{enumerate}

\vspace{-3mm}

\subsection{Retrieval and constraint checking}
Given a user query $q$, the system proceeds in two stages:

\vspace{-1mm}

\paragraph{Stage 1: Candidate generation (structural filtering).}
Using $C_s$, we query the graph to retrieve a top-$k$ set of candidates:
\[
\mathcal{P}_k = \mathrm{TopK}\big(\{p \in \mathcal{P} \mid p \models C_s\}\big),
\]
where $p \models C_s$ denotes that $p$ satisfies \emph{all} structural constraints.



\textit{Stage 2: Evidence-based semantic filtering (within retrieved candidates).}
We filter retrieved candidates $\mathcal{P}_k$ with semantic constraints:
\[
\mathcal{P}^{\text{sem}}_k \;=\; \{\, p \in \mathcal{P}_k \;|\; \mathrm{check}(p, C_m)=1 \,\}.
\]
Here, $\mathrm{check}(p,C_m)$ is a semantic constraint checker computed from evidence retrieved from the
candidate's text fields. If $\mathcal{P}^{\text{sem}}_k=\emptyset$, we fetch the next top-$k$ candidates
and repeat \emph{Stage 2}.




\vspace{-2mm}

\section{Experimental Setup}
\label{sec:setup}

\subsection{Dataset curation}
We evaluate on a curated food pantry directory containing \(\sim\)800 pantries with structured fields (name, address, city, county, ZIP code, phone, hours) and unstructured fields (eligibility). We use a publicly available Kansas directory\footnotemark{}\footnotetext{Kansas Food Source. Find Food in the Sunflower State. \url{https://kansasfoodsource.org/}} because it provides statewide coverage and access to verified pantry listings. This pipeline applies to any public-service directory and can be extended to other domains (e.g., clinics, shelters, transportation services). For evaluation, this directory is assumed to be the only desired source of information for the purpose of our study, and any suggestions not included in this dataset are counted as out-of-dataset hallucinations.

\vspace{-1mm}

\subsection{Knowledge graph variants}
To study how knowledge representation via knowledge graphs (KGs) affects retrieval, we compare the following:
\begin{itemize}
    \item \textbf{KG-0 (No-KG Baseline):} text-only indexing over pantries.
    \item \textbf{KG-1 (Location KG):} a location-modeled graph linking \texttt{Pantry} nodes to \texttt{City}/\texttt{County}/\texttt{Zipcode}.
    \item \textbf{KG-2 (Hours KG):} a graph linking \texttt{Pantry} to \texttt{Hours}.
    \item \textbf{KG-3 (Location + Hours KG):} a combined graph including both location and hours relations (KG-1 + KG-2).
\end{itemize}

\vspace{-3mm}

\subsection{Query benchmark and ground truth}
We construct a synthetic benchmark of 1{,}000 user queries to reflect common public-service information needs. Queries are grouped into five families (see Table~\ref{tab:query_families}) covering: (i) location-only requests (city/county/ZIP), (ii) eligibility questions (ID), (iii) open-hours queries (day/time), (iv) recall-style queries that mention a specific pantry name, and (v) queries combining location with eligibility and/or hours constraints.
For evaluation, we derive directory-grounded ``gold'' answers by (1) applying strict structural filtering over the pantry directory for the structural constraints in the query and (2) returning the top candidates as the reference context.

\vspace{-3mm}

\begin{table}[htbp]
\centering
\caption{Distribution of benchmarked queries across families.}
\label{tab:query_families}
\vspace{-3mm}
\setlength{\tabcolsep}{3.5pt}
\renewcommand{\arraystretch}{1.1}
\begin{tabular}{lcccccc}
\toprule
\textbf{Family} &
\textbf{Location} &
\textbf{Eligibility} &
\textbf{Hours} &
\textbf{Recall} &
\textbf{Combined} \\
\midrule
\textbf{Queries} & 500 & 200 & 200 & 10 & 90 \\
\bottomrule
\end{tabular}
\vspace{-7mm}
\end{table}


\subsection{Models evaluated}

We evaluate 15 open-source instruction-tuned LLMs across diverse model families like Qwen \cite{yang2025qwen3}, Llama \cite{grattafiori2024llama}, Gemma \cite{team2024gemma}, Phi \cite{abdin2024phi}, DeepSeek R1 Distill \cite{guo2025deepseek}, spanning small (\(\leq\)3B), medium (4--8B), and large (\(>\)8B) parameter size groups. We also include one stronger proprietary baseline, GPT 5.5 \cite{singh2025openai}. All models are tested with the same queries mentioned in Table \ref{tab:query_families}. 

\vspace{-4mm}

\subsection{Evaluation metrics}

\noindent \textbf{Retrieval quality.} We report Precision@k and Recall@k (with $k\in\{3,5\}$) against the gold candidate set for each query. We also report F1@k as the harmonic mean of Precision@k and Recall@k~\cite{manning2008introduction}.

\noindent \textbf{Constraint satisfaction.} Fraction of queries where all explicitly stated constraints are satisfied by the returned recommendations.

\noindent \textbf{Hallucination rate.} Fraction of queries where the response includes at least one recommended pantry not present in the directory.

\noindent \textbf{Semantic similarity.} We measure cosine similarity between fixed sentence embeddings of the model-generated responses and the directory-grounded gold standard responses \cite{hliaoutakis2009information}. Higher values indicate closer semantic alignment.

\vspace{-3mm}

\section{Results and Discussion}
\label{sec:results}

\begin{table*}[t]
\centering
\small
\setlength{\tabcolsep}{4pt}
\begin{tabular}{ll|cccc|cccc|cccc}
\toprule
Model Group & Model Name &
\multicolumn{4}{c|}{Constraint Satisfaction (\%)} &
\multicolumn{4}{c|}{Hallucination Rate (\%)} &
\multicolumn{4}{c}{Semantic Similarity} \\
& & KG-0 & KG-1 & KG-2 & KG-3 & KG-0 & KG-1 & KG-2 & KG-3 & KG-0 & KG-1 & KG-2 & KG-3 \\
\midrule
\textbf{Proprietary}
& GPT-5.5
& 72.18 & 92.06 & 93.21 & \textbf{97.94}
& 5.71 & 1.63 & 1.12 & \textbf{0.61}
& 0.75 & 0.84 & 0.85 & \textbf{0.87} \\
\midrule
\textbf{Large}
& Llama 3.3 70B Instruct
& 73.58 & 90.08 & \textbf{98.11} & 97.63
& 5.41 & 1.92 & \textbf{0.92} & 0.97
& 0.74 & 0.83 & \textbf{0.86} & 0.85 \\
& DeepSeek R1 Distill Qwen (32B)
& 72.33 & 89.04 & 90.13 & \textbf{97.02}
& 5.96 & 1.88 & 1.21 & \textbf{0.96}
& 0.73 & 0.81 & 0.82 & \textbf{0.84} \\
& Gemma 4 31B Instruct
& 71.09 & \textbf{95.44} & 94.67 & 95.18
& 6.10 & \textbf{0.99} & 1.06 & 1.01
& 0.72 & \textbf{0.83} & 0.82 & 0.83 \\
& DeepSeek R1 Distill Qwen (14B)
& 70.12 & 86.03 & 87.44 & \textbf{94.19}
& 6.71 & 2.24 & 1.58 & \textbf{1.01}
& 0.71 & 0.78 & 0.79 & \textbf{0.81} \\
\midrule
\textbf{Medium}
& Llama 3.1 8B Instruct
& 66.08 & 83.52 & 85.11 & \textbf{91.33}
& 9.01 & 2.92 & 2.11 & \textbf{1.02}
& 0.67 & 0.75 & 0.76 & \textbf{0.78} \\
& DeepSeek R1 Distill Llama (8B)
& 66.57 & 84.01 & \textbf{92.08} & 91.22
& 8.63 & 2.83 & \textbf{0.98} & 1.05
& 0.68 & 0.76 & \textbf{0.79} & 0.78 \\
& Qwen 3 8B
& 65.42 & 82.10 & \textbf{90.06} & 88.83
& 9.43 & 3.03 & \textbf{1.04} & 1.18
& 0.67 & 0.74 & \textbf{0.77} & 0.76 \\
& Qwen 2.5 7B Instruct
& 64.89 & 81.14 & 82.05 & \textbf{89.27}
& 9.88 & 3.39 & 2.94 & \textbf{1.91}
& 0.66 & 0.73 & 0.74 & \textbf{0.76} \\
& Qwen 3 4B
& 61.77 & 78.03 & 80.11 & \textbf{86.48}
& 11.52 & 4.26 & 3.57 & \textbf{2.06}
& 0.65 & 0.72 & 0.73 & \textbf{0.75} \\
& Phi 3.5 Mini Instruct (4B)
& 62.21 & \textbf{86.02} & 84.91 & 85.40
& 10.83 & \textbf{2.17} & 2.35 & 2.28
& 0.65 & \textbf{0.75} & 0.74 & 0.74 \\
\midrule
\textbf{Small}
& Qwen 2.5 3B Instruct
& 58.66 & 74.08 & 76.77 & \textbf{83.29}
& 14.31 & 5.43 & 4.61 & \textbf{3.22}
& 0.62 & 0.69 & 0.70 & \textbf{0.72} \\
& Llama 3.2 3B Instruct
& 57.84 & 75.46 & \textbf{84.07} & 82.62
& 14.89 & 5.11 & \textbf{2.88} & 3.02
& 0.61 & 0.70 & \textbf{0.73} & 0.72 \\
& Gemma 3 1B Instruct
& 56.03 & \textbf{80.44} & 78.12 & 79.38
& 15.67 & \textbf{3.91} & 4.10 & 4.02
& 0.60 & \textbf{0.71} & 0.70 & 0.70 \\
& Llama 3.2 1B Instruct
& 55.27 & 71.31 & \textbf{79.06} & 77.82
& 16.44 & 6.52 & \textbf{4.02} & 4.28
& 0.60 & 0.66 & \textbf{0.70} & 0.69 \\
& Qwen 2.5 0.5B Instruct
& 52.41 & 69.22 & 72.08 & \textbf{78.11}
& 18.09 & 7.71 & 6.88 & \textbf{5.39}
& 0.58 & 0.65 & 0.66 & \textbf{0.69} \\
\midrule
\textbf{Average across} & \textbf{all models} 
& 64.15 & 82.37 & 85.50 & \textbf{88.48}
& 10.54 & 3.50 & 2.59 & \textbf{2.12}
& 0.67 & 0.75 & 0.76 & \textbf{0.77} \\
\bottomrule
\end{tabular}
\caption{
End-to-end results across LLMs and KG variants. Bold indicates the best-performing KG per metric for each model.
}
\label{tab:merged}
\vspace{-3mm}
\end{table*}


\begin{figure*}[t]
\centering
\vspace{-5mm}
\begin{tikzpicture}
\begin{axis}[
width=0.29\textwidth,
height=4.2cm,
xmin=0,xmax=3,
ymin=50,ymax=100,
xtick={0,1,2,3},
xticklabels={KG-0,KG-1,KG-2,KG-3},
ylabel={Constraint Satisfaction (\%)},
title={(a)}
]
\foreach \a/\b/\c/\d in {
52.41/69.22/72.08/78.11,
55.27/71.31/79.06/77.82,
56.03/80.44/78.12/79.38,
58.66/74.08/76.77/83.29,
57.84/75.46/84.07/82.62}
{
\addplot[color=smallc,line width=0.8pt,mark=|]
coordinates {(0,\a)(1,\b)(2,\c)(3,\d)};
}
\foreach \a/\b/\c/\d in {
61.77/78.03/80.11/86.48,
62.21/86.02/84.91/85.40,
64.89/81.14/82.05/89.27,
65.42/82.10/90.06/88.83,
66.08/83.52/85.11/91.33,
66.57/84.01/92.08/91.22}
{
\addplot[color=mediumc,line width=1pt,mark=|]
coordinates {(0,\a)(1,\b)(2,\c)(3,\d)};
}
\foreach \a/\b/\c/\d in {
70.12/86.03/87.44/94.19,
71.09/95.44/94.67/95.18,
72.33/89.04/90.13/97.02,
73.58/90.08/98.11/97.63}
{
\addplot[color=largec,line width=1.2pt,mark=|]
coordinates {(0,\a)(1,\b)(2,\c)(3,\d)};
}
\addplot[color=propc,line width=1.5pt,mark=|]
coordinates {(0,72.18)(1,92.06)(2,93.21)(3,97.94)};
\end{axis}
\begin{axis}[
at={(5.4cm,0)},
anchor=south west,
width=0.29\textwidth,
height=4.2cm,
xmin=0,xmax=3,
ymin=0,ymax=20,
xtick={0,1,2,3},
xticklabels={KG-0,KG-1,KG-2,KG-3},
ylabel={Hallucination Rate (\%)},
title={(b)}
]
\foreach \a/\b/\c/\d in {
18.09/7.71/6.88/5.39,
16.44/6.52/4.02/4.28,
15.67/3.91/4.10/4.02,
14.31/5.43/4.61/3.22,
14.89/5.11/2.88/3.02}
{
\addplot[color=smallc,line width=0.8pt,mark=|]
coordinates {(0,\a)(1,\b)(2,\c)(3,\d)};
}
\foreach \a/\b/\c/\d in {
11.52/4.26/3.57/2.06,
10.83/2.17/2.35/2.28,
9.88/3.39/2.94/1.91,
9.43/3.03/1.04/1.18,
9.01/2.92/2.11/1.02,
8.63/2.83/0.98/1.05}
{
\addplot[color=mediumc,line width=1pt,mark=|]
coordinates {(0,\a)(1,\b)(2,\c)(3,\d)};
}
\foreach \a/\b/\c/\d in {
6.71/2.24/1.58/1.01,
6.10/0.99/1.06/1.01,
5.96/1.88/1.21/0.96,
5.41/1.92/0.92/0.97}
{
\addplot[color=largec,line width=1.2pt,mark=|]
coordinates {(0,\a)(1,\b)(2,\c)(3,\d)};
}
\addplot[color=propc,line width=1.5pt,mark=|]
coordinates {(0,5.71)(1,1.63)(2,1.12)(3,0.61)};
\end{axis}
\begin{axis}[
at={(10.8cm,0)},
anchor=south west,
width=0.29\textwidth,
height=4.2cm,
xmin=0,xmax=3,
ymin=0.55,ymax=0.90,
xtick={0,1,2,3},
xticklabels={KG-0,KG-1,KG-2,KG-3},
ylabel={Semantic Similarity},
title={(c)}
]
\foreach \a/\b/\c/\d in {
0.58/0.65/0.66/0.69,
0.60/0.66/0.70/0.69,
0.60/0.71/0.70/0.70,
0.62/0.69/0.70/0.72,
0.61/0.70/0.73/0.72}
{
\addplot[color=smallc,line width=0.8pt,mark=|]
coordinates {(0,\a)(1,\b)(2,\c)(3,\d)};
}
\foreach \a/\b/\c/\d in {
0.65/0.72/0.73/0.75,
0.65/0.75/0.74/0.74,
0.66/0.73/0.74/0.76,
0.67/0.74/0.77/0.76,
0.67/0.75/0.76/0.78,
0.68/0.76/0.79/0.78}
{
\addplot[color=mediumc,line width=1pt,mark=|]
coordinates {(0,\a)(1,\b)(2,\c)(3,\d)};
}
\foreach \a/\b/\c/\d in {
0.71/0.78/0.79/0.81,
0.72/0.83/0.82/0.83,
0.73/0.81/0.82/0.84,
0.74/0.83/0.86/0.85}
{
\addplot[color=largec,line width=1.2pt,mark=|]
coordinates {(0,\a)(1,\b)(2,\c)(3,\d)};
}
\addplot[color=propc,line width=1.5pt,mark=|]
coordinates {(0,0.75)(1,0.84)(2,0.85)(3,0.87)};
\end{axis}
\end{tikzpicture}
\vspace{-5mm}
\caption{
Performance trends of constraint satisfaction, hallucination rate, and semantic similarity across KG variants. Darker lines indicate larger models.
}
\vspace{-5mm}
\label{fig:kg}
\end{figure*}
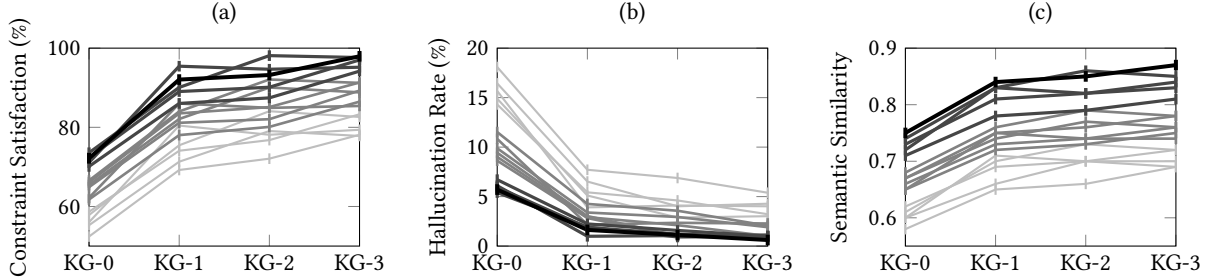


\subsection{Retrieval performance across KG variants}
Table~\ref{tab:retrieval} reports retrieval quality for each KG variant at $k\in\{3,5\}$ using Precision@k, Recall@k, and F1@k. These metrics evaluate only the retrieval, i.e., the quality of the top-$k$ candidate set returned before any LLM response generation. Overall, KG-based representations substantially improve retrieval compared to the no-KG baseline. At $k=3$, KG-3 achieves the strongest retrieval across all reported metrics, reaching P@3=0.87, R@3=0.91, and F1@3=0.86 respectively. A similar pattern holds at $k=5$, where KG-3 remains best, while KG-0 remains the weakest baseline. These results suggest that combining location and hours in KG-3 yields the most accurate retrieved candidate sets prior to downstream constraint checking and response generation via LLM.


\begin{table}[t]
\centering
\caption{Retrieval quality for each KG variant at $k{=}3, 5$, reported via Precision@k (P@k), Recall@k (R@k), and F1@k. Best values per column are marked in bold.}
\vspace{-3mm}
\label{tab:retrieval}
\setlength{\tabcolsep}{3pt}
\renewcommand{\arraystretch}{1.1}
\small
\begin{tabular}{lcccccc}
\toprule
\textbf{KG Variant} & \textbf{P@3} & \textbf{R@3} & \textbf{F1@3} & \textbf{P@5} & \textbf{R@5} & \textbf{F1@5}\\
\midrule
KG-0 (No-KG)          & 0.38 & 0.40 & 0.39 & 0.30 & 0.44 & 0.36 \\
KG-1 (Location)       & 0.70 & 0.72 & 0.71 & 0.62 & 0.75 & 0.68 \\
KG-2 (Hours)          & 0.66 & 0.63 & 0.64 & 0.56 & 0.70 & 0.62 \\
KG-3 (Location + Hours) & \textbf{0.87} & \textbf{0.91} & \textbf{0.86} & \textbf{0.77} & \textbf{0.93} & \textbf{0.84} \\
\bottomrule
\vspace{-9mm}
\end{tabular}
\end{table}

\vspace{-5mm}

\subsection{Reliability improves with stronger retrieval}
Table~\ref{tab:merged} and Figure~\ref{fig:kg} (a,b,c) show a consistent reliability trend as the KG representation becomes more structured. Averaged across all models, constraint satisfaction increases from 64.15\% (KG-0) to 88.48\% (KG-3) while semantic similarity also improves (0.67 to 0.77). The upward trajectories in Figure~\ref{fig:kg} (a,c) indicate that retrieval-centric improvements dominate the end-to-end behavior across model families, supporting the premise that candidate quality is a primary bottleneck in directory-based public-service QA.

\vspace{-3mm}

\subsection{Reducing out-of-dataset hallucinations}
A key benefit of stronger retrieval is the reduction in out-of-dataset recommendations. Table~\ref{tab:merged} shows that the hallucination rate drops from 10.54\% under KG-0 to 2.12\% under KG-3 on average, and Figure~\ref{fig:kg} (b) confirms a sharp decline from KG-0 to KG-1 that continues through KG-3. This supports the interpretation that when the LLM is restricted to a verified candidate set retrieved under explicit constraints, it becomes less likely to suggest outside the directory.

\vspace{-3mm}

\subsection{Retrieval reduces sensitivity to model choice}
Figure~\ref{fig:kg} (a,b,c) highlights that model-to-model variance is the largest under KG-0 and shrinks under stronger KG variants. In particular, as retrieval improves (KG-1 to KG-3), performance curves across models become more tightly clustered, suggesting that the LLM’s role shifts toward shallow reasoning over retrieved evidence. This pattern aligns with our central claim: in public-service directory settings, stronger retrieval reduces sensitivity to model choice and enables more robust performance even with smaller models.

\vspace{-3mm}

\section{Conclusion}
\label{sec:conclusion}

This paper investigates whether retrieval quality determines recommendation reliability in public service conversational agents. To answer this, we propose TRACE, a constraint-aware retrieval framework grounded in knowledge graph representations of public service directories. Our experiments demonstrate that improving retrieval consistently improves end-to-end recommendation reliability, and that stronger retrieval stabilizes performance across diverse LLMs. Together, these findings establish retrieval as the critical bottleneck in public service conversational agents. 





\bibliographystyle{ACM-Reference-Format}
\bibliography{software}


\end{document}